\documentclass[10pt, conference]{IEEEtran}
\pdfoutput=1

\usepackage[cmex10]{amsmath}
\usepackage{amssymb}

\usepackage{cleveref}
\usepackage[pdftex]{graphicx}
\usepackage{tabularx}
\usepackage[font=footnotesize]{subfig}

\title{Strategic Control of Proximity Relationships\\ in Heterogeneous Search and Rescue Teams}

\author{\IEEEauthorblockN{Eduardo Feo Flushing, Luca M. Gambardella, Gianni A. Di Caro}
\IEEEauthorblockA{Dalle Molle Institute for Artificial Intelligence (IDSIA), Lugano, Switzerland\\
Email: \{eduardo,luca,gianni\}@idsia.ch}}

\begin{document}
\newcommand{\Agt}{\mathcal{A} } 
\newcommand{\XA}{\mathcal{X}^{p}} 
\newcommand{\XB}{\bar{\mathcal{X}} }
\newcommand{\X}{\mathcal{X} }
\newcommand{\Aloc}{\Gamma}
\newcommand{\Bloc}{\mathcal{C}}
\newcommand{\rLoc}{\theta}
\newcommand{\ene}{\lambda}
\newcommand{\ENE}{\bar{\lambda}}
\newcommand{\enebudget}{\Lambda}
\newcommand{\Act}{\mathcal{O}}
\newcommand{\Loc}{\mathcal{L}}

\newcommand{\expl}{\varphi}
\newcommand{\EXPL}{\Phi}
\newcommand{\sd}{\varsigma}
\newcommand{\YT}{Y^{T}}
 \newcommand{\Tau}{\mathrm{T}}
 \newcommand{\YCOL}{Y^{\tau}}
 \newcommand{\YTWO}{\mathcal{Y}}
 \newcommand{\ytwo}{\psi}
 \newcommand{\ytwomin}{\Psi}
 \newcommand{\dev}{\gamma}
 \newcommand{\dAloc}{\Aloc^{\gamma}}
 \newcommand{\Atype}{\Omega}

\newcommand{\MinDist}{\Theta^{t}}
\newcommand{\sMinDist}{\hat{\Theta}^{t}}
\newcommand{\hMinDist}{\bar{\Theta}^{t}}
\newcommand{\bMinDist}{\breve{\Theta}^{t}}

\newcommand{\MaxDist}{\Psi^{t}}
\newcommand{\hMaxDist}{\bar{\Psi}^{t}}
\newcommand{\bMaxDist}{\breve{\Psi}^{t}}
\newcommand{\sMaxDist}{\hat{\Psi}^{t}}

\newcommand{\MT}{\mathit{T}}
\newcommand{\mT}{1}
\newcommand{\degree}{^{\circ}}
\newcommand{\explup}{\bar{\expl}}
\newcommand{\Tra}{\mathcal{T}}
\newcommand{\realexpl}{\hat{\expl}}
\newcommand{\groupmincc}{\Upsilon_{min}}
\newcommand{\groupmaxcc}{\Upsilon_{max}}
\newcommand{\minHAS}{\Pi}
\newcommand{\shadowd}{\hat{d}}
\newcommand{\cellprop}{p^{*}}
\newcommand{\elog}{\mathcal{L}}
\newcommand{\schedule}{\mathcal{S}}
\newcommand{\colaN}{\bar{\mathcal{A}}}
\newcommand{\colaR}{\tau}
\newcommand{\ccon}{\beta}
\newcommand{\txr}{\psi_{\text{r}}}
  \newcommand{\yt}{y^{t}}
\newtheorem{defi}{Definition}
\newcommand{\dt}{d^{t}}
\newcommand{\reccon}{\gamma}
\newcommand{\CCON}{\mathit{I}}
\newcommand{\RECCON}{\mathit{R}}

\maketitle
\thispagestyle{empty}
\pagestyle{empty}

\begin{abstract}
In the context of search and rescue, we consider the problem of mission planning for heterogeneous
teams that can include human, robotic, and animal agents. The problem is tackled using a mixed
integer mathematical programming formulation that jointly determines the path and the activity
scheduling of each agent in the team. Based on the mathematical formulation, we propose the use of
soft constraints and penalties that allow the flexible strategic control of spatio-temporal relations
among the search trajectories of the agents.  In this way, we can enable the mission planner to
obtain solutions that maximize the area coverage and, at the same time, control the spatial
proximity among the agents (e.g., to minimize mutual task interference, or to promote local
cooperation and data sharing). Through simulation experiments, we show the application of the
strategic framework considering a number of scenarios of interest for real-world search and rescue
missions.
\end{abstract}

\section{Introduction}

%
The increasing availability of new technologies that can be profitably employed in search and
rescue (SAR) missions, including aerial autonomous robotic platforms, GPS-enabled devices for
human rescuers, and augmenting technologies for tracking and controlling animal agents (e.g.,
dogs), has raised the need for devising new ways of planning and executing SAR missions.
%
In fact, in the resulting {\em heterogeneous teams}, some agents are more capable to perform certain
tasks than others, leading to the complex problem of effectively integrating the different
capabilities and assigning roles and
responsibilities during the mission.
Thus, {\em mission planning} for heterogeneous teams needs to jointly decide and coordinate
the activities of the agents in order to perform the search tasks as efficiently as possible,
exploiting at the best the individual capabilities and potential mutual synergies.

In previous work~\cite{FeoGamDic12:rosin,AugFeo12:tr-idsia:05-12-n} 
we formulated the SAR mission
planning as a mixed-integer linear
programming (MIP) optimization problem
and we focused on 
SAR in {\em large wilderness areas} (WiSAR).
%
The objective was to jointly define, for all agents in a heterogeneous SAR team, the search
trajectories and the activity scheduling that maximize the {\em coverage} of the area.  Resulting
trajectories consist of sequences of adjacent environment sectors, while activity scheduling
specifies the amount of time that each agent should spend searching inside each sector
along its trajectory.
%

In practice, the maximization of the overall area coverage, as considered before, 
might not be the only representing factor of efficiency for mission performance. 
Other important aspects, common to SAR missions, need to be accounted for 
in order to be able to deal with the complexity of the real-world and to address the needs and the
possibilities offered by the presence of a heterogeneous team.
Among these aspects, the management of {\em agents' interactions} in terms of {\em spatio-temporal
relations}, represents a  
challenging issue to the mission planner, and
it may include the following:
(i)~provisioning of topological connectivity to support {\em
  wireless communications} (e.g., to stream mission updates to a control center);
(ii)~{\em boosting cooperation} between the agents,
(e.g.,  if a human rescuer and a robot are concurrently exploring 
close-by or overlapping areas,
the robot could augment the rescuer's view by serving requests for the
real-time video streaming of areas selected by the human);
(iii)~{\em increasing safety of agents} (e.g., at night time, the human agents 
performing a search in the wilderness might be required to
stay relatively close to each other for personal safety reasons),
and (iv)  {\em minimizing negative interferences} 
(e.g., different air-scent dogs should be sent out
over different, far away areas to avoid that they disturb each other).

In this work, we extend our previous approach in order to compose a general strategic framework that
addresses the issues discussed above.  
The framework is based on the promotion or enforcement
of spatio-temporal relations among the search trajectories of the agents.  

\section{Related Work}
Broadly speaking, the approaches for
the planning of SAR missions can be
divided into two categories: probabilistic and non-probabilistic. 
%
When a probabilistic framework is considered, 
the mission assignment attempts to optimize 
the search by maximizing detection
probability~\cite{Lin2009},
minimizing the expected time for
target detection~\cite{Lau2006}, 
or by or number of detections~\cite{Chung2009}.  
Although we do not explicitly formulate the
problem in probabilistic terms, 
we use the related notion of {\em exploration demands}: 
bounded range variables used to prioritize parts of the area according to the
expected amount of exploration they require.
Among these works, only a few have considered the inclusion of spatio-temporal constraints,
and aimed at specific application scenarios such as network connectivity~\cite{HolSin12}, 
and coalition avoidance~\cite{GanFitSuk12}.

Outside the probabilistic context, the problem of multi-agent mission planning 
is the subject of a
large amount of research in the fields of operations research (OR) and multi-robot systems.  In
particular, as {\em vehicle routing} (in OR) and {\em multi-robot task allocation problems}, which
are indeed two large families of problems, sharing the objective of efficiently allocating spatially
distributed tasks to a team of agents.

Vehicle routing problems (VRPs) have been the starting point 
of several planning models for complex missions,
such as military operations~\cite{MufBatNag12}, and space
exploration~\cite{AhnDewGenKla12}.
Similarly, we formulate the mission planning as a variant of VRPs 
that introduces several
aspects motivated by SAR applications.
Among these we have the integration of scheduling decisions into the routing problem, and rewards
dependent upon decidable service time.
In particular, we consider the definition of spatio-temporal relations among the trajectories of the agents,
which is a case of strong inter-dependencies between vehicles/agents.  
All these aspects represent new challenges for VRPs~\cite{Dre12} 
which have not previously been studied
in-depth, and are of practical relevance.

In relation to research in robotics, the class of problems closest to ours 
is that of {\em multi-robot task allocation}, which 
is the problem of determining 
which robots should execute which tasks in order to achieve the 
overall system goals. 
According to the categorization of Gerkey and Mataric~\cite{GerMat04}, our planning problem covers
both {\em single-task robots} (ST) and {\em multi-task robots} (MT), and considers {\em single-robot
  tasks} (SR), for a {\em time-extended assignment} (TA).
  
Within this domain, only a few studies have considered 
scenarios with inter-dependencies between agents' plans, 
in which the plan of single agent may affect the feasibility and/or utility
of other agents' plans \cite{KorKanBroSteDia12}.
Most of these have been limited to
the provisioning of ad hoc communication in a team of agents executing a mission \cite{HolSin12,FeoKudGamDic13:ssrr}.
The strategic decision framework presented in this work 
involves the definition of a powerful, and generic, class of spatio-temporal directives, 
capable of {\em introducing} explicit {\em dependencies in space and time} 
among the plans of individual agents.
Altogether, the planning model and the definition of directives, 
constitute a strategic decision framework that allows the management 
of proximity relationships among the agents,
thus making our model 
much closer to the real-world than previous ones,
and providing truly flexible tools for the mission commander.

\section{System Model}
 \label{sec:model}
In this section we present the way we modeled the reference WiSAR scenario, which is directly
derived from our previous work~\cite{FeoGamDic12:rosin,AugFeo12:tr-idsia:05-12-n}.
As a first step, the search area is discretized into a set of squared {\em
  environment cells}.
The discretization ease the computation and the assignment of
properties to the local environment, and helps to evaluate the status of the search and to measure
the performance of mission plans.
Without losing generality, in the following a uniform cell grid decomposition is considered, and the
cells' set is indicated by $\Bloc = \lbrace c_1,\ldots,c_n \rbrace$.  
Environment cells also serve as a
means of evaluating mission status in terms of coverage.  The {\em coverage map} $C_m : \Bloc
\mapsto [0,1]$, relates cells to numerical values representing the amount of coverage currently required, 
or, in other words, the {\em residual need of exploration} of each cell.

Based on the above cell definitions, the area is further decomposed into 
{\em sectors}, that
is, clusters of contiguous environment cells (i.e., a subset of $\Bloc$). 
Sectors are needed to account for the different sensory-motor capabilities of the different agents, and for the purpose of
efficient search, since performing mission planning at the resolution of individual environment
cells may become both impractical and computationally infeasible, if cells are very small. 


%
%
Defining the sector boundaries requires a careful analysis, and ideally, knowledge of the region.
In case of heterogeneous team of agents, the definition of possible sectors must be done taking into
account the agents' capabilities and terrain conditions in order to properly pair them.  
In the following, the set of heterogeneous agents composing the WiSAR team is represented by $\Agt$,
and $\Aloc_k$ denotes the set of {\em assignable} sectors to agent $k \in \Agt$.
Note that plans of different agents may be specified in terms of totally different sets of sectors
(e.g., plans of aerial robots may involve search tasks over areas much larger than those
considered in plans of ground resources), 
in this way we account for their different characteristics of mobility and sensing.
Additionally, there are no restrictions imposed on the elements of $\Aloc_k$.
For instance, $\Aloc_k$ may be composed of sectors of different shapes and sizes,
which may overlap between themselves. 
For sake of simplicity, we assume that sets $\Aloc_k$, for each $k \in \Agt$,
are provided as input, 
specifically as a list of geographically delimited regions (i.e., polygons on the earth's surface).
%

%
Once the sectors has been defined, then the issue becomes that of allocating the resources
inside the sectors and deciding when, from whom, and how much effort each sector will receive. 
This is accomplished by the specification of {\em agent plans}, which are defined in terms of {\em search
  tasks}: dispatching the agents to sectors with the objective of carrying out exploration
activities for a certain amount of time. A {\em global mission plan} consists of sequences of search
tasks to be executed one after the other by each one of the agents.  
Search tasks are represented by
$\langle L, t_{start}, t_{end} \rangle$, where $L\subseteq \Bloc$ is a sector, and $t_{start}$ and
$t_{end}$ are start and end times for a search task inside sector $L$.  
For simplicity, and without
losing generality, the whole mission time is discretized into {\em mission intervals} of
length equal to $\Delta_t$ seconds. That is, $\Delta_t$ is the common {\em time unit} for the starting,
ending, and duration of all search tasks.

For planning purposes, we assume that the traveling times between tasks 
are implicitly taken into account within the time allocated to perform the tasks.
Clearly, traveling across a sector can be performed while searching, and, in fact, 
it is typically done during the search.
Under this consideration, we remark that the set of sectors of each agent 
must be congruent to the definition of the mission interval length $\Delta_t$.
Given that this interval represents the minimum duration of any task assigned to 
any agent, its length must be enough to enable each agent to traverse the sector 
corresponding to its current task and ensure the timely arrival to the sector assigned for the next task.
Therefore, mission plans should be designed taking into account that consecutive tasks
belonging to a plan must involve sectors that are physically close, if not adjacent.
To this end, we define a {\em traversability graph} $G_k=(\Aloc_k, E_k)$  
for each agent $k \in \Agt$, 
where $E_k$ contains an edge $(i,j)$ 
if a task at sector $j$ 
can be scheduled right after a task at sector $i$.

\begin{figure}[!]
\centering
\includegraphics[width=0.5\columnwidth]{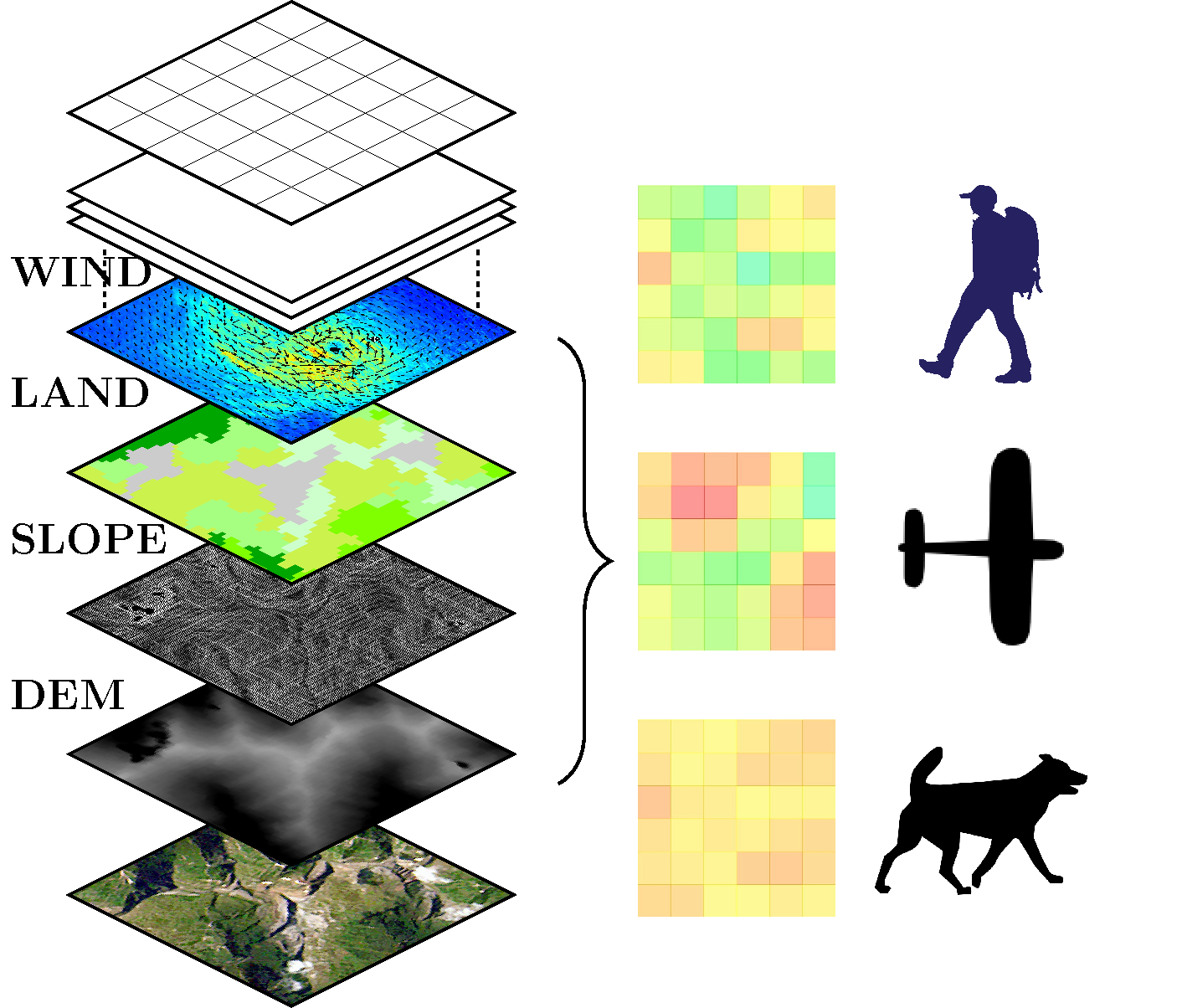}
\caption{Example of estimation search efficacy using GIS.\vspace*{-1ex}}
\label{fig:gis}
\end{figure}

In order to compute efficient joint mission plans, the planner must explicitly take into account the
fact that different agents may show different levels of performance accomplishing a task in
the same portion of environment, due to their heterogeneous skills as well as to the effect of local
conditions (e.g., unmanned aerial vehicles operating in densely vegetated areas may not be able to
effectively detect targets on the ground using vision sensors, humans walking up a steep hill might
move at a considerably slow pace).  One way to accomplish this is to extract relevant environment
properties from spatial data provided by {\em geographical information systems} (GIS) and to define
procedures for estimating the {\em expected search performance} (also termed {\em search efficacy}
in the following), for each single agent and for each different portion of
the environment~\cite{FeoGamDic12:rosin}, as shown in Figure~\ref{fig:gis}.
In the following, we assume that the search efficacy is
specified by the \emph{coverage rate}, $\realexpl_k$.  
When an agent $k$ performs a 
search task inside cell $c$ for $t$ seconds, it provides the
$100\cdot\realexpl_k(c)t$ percent of its coverage.  
The coverage rate can also be used to update $C_m$,
by tracking the movements of each agent and adding up the amount of time spent inside each cell.

Since mission planning is done in terms of assignments of sectors to agents, the expected search
performance $\expl_k(L,c)$ serves as a tool to {\em estimate} the coverage that will be done inside
cell $c$ by the activities of agent $k$ inside sector $L$ for $\Delta_t$ seconds.  In other words,
$\expl_k$ provides an estimation of the way the effort of $k$ will be split among cells composing a
sector.  Without losing generality, we assume that the time assigned to a sector
is uniformly distributed among all its composing cells.  For $c \in \Bloc$, $k \in \Agt$, and
$L \in \Aloc_k$:
\vspace*{-2ex}

{\small
\begin{gather}
\expl_k(L,c) = \begin{cases} \realexpl_k \left(c \right) \left(\dfrac{\Delta_t}{\vert L \vert}\right) & \text{if } c \in L \\ 0 & \text{otherwise}\end{cases}
\end{gather}
}

\begin{figure}[t]
\centering
\includegraphics[width=0.68\columnwidth]{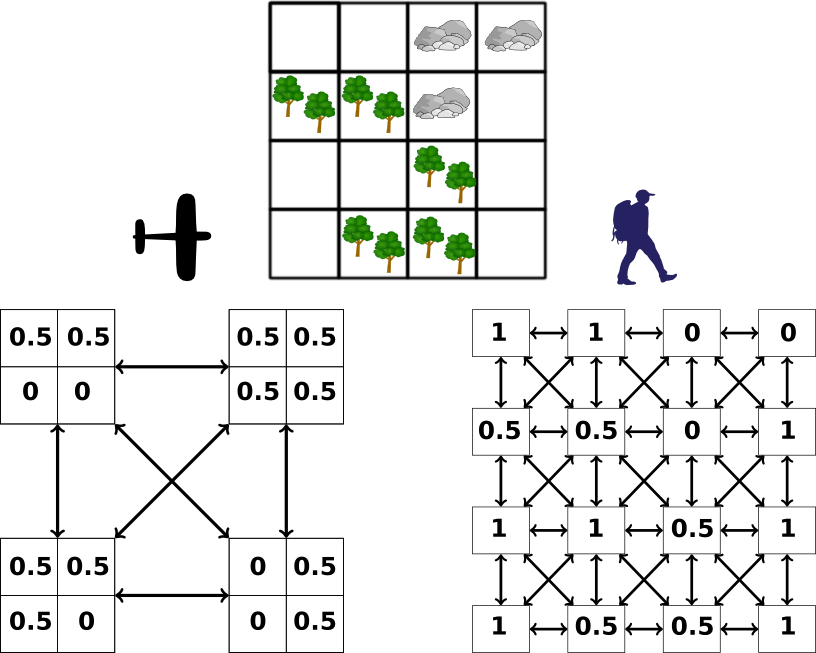}
\caption{Sector layouts for two agents. \vspace*{-0ex}}
\label{fig:ex_sectors}
\end{figure}

Fig.~\ref{fig:ex_sectors} represents an example of sector decomposition,
together with a definition of $\expl$. In the figure, the area has been decomposed into
16 cells, as shown in the square grid at the top. Two different sector
layouts are defined, one for a human rescuer and the other for an aerial vehicle.
The layout on the right defines 16 sectors of same size as cells, thus a one-to-one correspondence
between cells and sectors. On the left, 4 sectors composed by 4 cells each are defined. 
The numbers inside the figure correspond to the estimated search performance $\expl_k(L,c)$ for agents
corresponding to each one of the layouts, taking into account the local characteristics of each cell 
(i.e., presence of vegetation or irregular terrain).

\section{Joint Mission Planning}
\label{sar-opt:sec}

We formulate mission planning as an {\em optimization problem}. 
The objective is to jointly define, for all agents in the SAR team, 
search trajectories and activity scheduling
seeking for the maximal efficiency in the mission performance.


Let $\Aloc^0_k \subseteq \Aloc_k$ be the  sectors 
which are accessible from the agent's initial position.
Given a limited time budget $\MT$ 
(e.g., mission's time span), 
a plan for an agent $k$ specifies an elementary path $p_k$ in $G_k$ 
(i.e., a sequence of sectors).
A path must start with a sector belonging to $\Aloc^0_k$, 
and does not necessarily include all sectors $\Aloc_k$,
due to the time constraints.
We denote the sectors {\em visited} in tasks assigned 
to agent $k$'s plan by  $v(p_k)$. 
The amount of time (expressed in $\Delta_t$ units) assigned to each sector in the 
plan is represented by the {\em schedule function}
$s_k: \Aloc_k \mapsto \mathbb{N}$,
with $s_k(v) > 0$ if $v \in v(p_k)$ and $s_k = 0$ otherwise.

The sum of the schedule of each agent must be equal
to the time budget $\MT$, that is $\sum_{v} s_k(v) = \MT$.
A solution to the mission planning problem consists of 
paths $p_k$ and schedules $s_k$
for all agents $k$ composing the team of rescuers $\Agt$,
and is denoted as $\mathcal{P} = \lbrace < p_k, s_k > \vert\ k \in \Agt \rbrace$.

The quality of a mission plan $\mathcal{P}$ 
in terms of {\em area coverage} is determined by the effect of 
agents' activities on the current status of the coverage map. The 
{\em coverage} of $\mathcal{P}$ is defined as follows, 
with  $C^{0}_m$ the initial coverage map:
\vspace*{-2ex}

{\small
\begin{gather}
\EXPL(\mathcal{P}) = \sum_{c \in \Bloc} \EXPL_c,\quad \text{where}, \label{eq:expp}
\end{gather}
}
\vspace*{-3ex}
{\small
\begin{gather}
\vspace*{-1ex}
\EXPL_c = \min \left( C^{0}_m(c), \displaystyle\sum_{k \in \Agt}\, \displaystyle\sum_{v \in v(p_k)} \expl_k(v,c)s_k(v) \right) \label{eq:expi}
\end{gather}
}
%

In other words, given the current coverage of cell $c$, the effect of plan $\mathcal{P}$ on the
coverage of $c$ ($\EXPL_c$) ranges from $0$ (no quantifiable effect) up to $C^{0}_m$ (completely
fulfilling the initial coverage requirements of $c$). Note that an environment cell can be affected
by the tasks of any number of agents.
%
\subsection{Mixed Integer Linear Model}
Making use of the notation introduced above, 
the mission planning is formalized as
a MIP problem.
%
The {\em decision variables} of the model are the following:\\
\hspace*{4pt}$x_{ijk}$: binary, equals 1 if agent $k$ traverses arc $(i,j) \in E_k  $;\\[0.3ex]
\hspace*{4pt}$y_{ik}$: binary, equals 1 if agent $k$ \emph{visits} sector $i \in \Aloc_k$;\\[0.3ex]
\hspace*{4pt}$\EXPL_{c}$: total coverage provided to cell $c \in \Bloc$ by all agents;\\[0.3ex]
\hspace*{4pt}$t_{ik}$: arrival time of agent $k$ at sector $i \in \Aloc_k$;\\[0.3ex]
\hspace*{4pt}$w_{ik}$: time spent by agent $k$ at sector $i \in \Aloc_k$.
\begin{figure}
  \begin{minipage}{1.0\columnwidth}
{\small
  \begin{gather}
\text{maximize } \displaystyle\sum_{c \in \Bloc } \EXPL_c  \label{mip:obj1}\\
\text{subject to} \nonumber \\
\displaystyle\sum_{(0,j) \in E_k} x_{0jk} = 1 \quad \forall k \in \Agt \label{mipc:ini}\\
\displaystyle\sum_{(j,0) \in E_k} x_{j0k} = 1 \quad \forall k \in \Agt \label{mipc:end}\\
\displaystyle\sum_{(i,j) \in E_k} x_{ijk} = \displaystyle\sum_{(j,i) \in E} x_{jik} = y_{jk} 
\quad \forall k \in \Agt,\  j \in \Aloc_k \label{mipc:flow}\\
\hspace*{-2ex}t_{ik} +  w_{ik} - t_{jk} \leq (1- x_{ijk})\MT\ \forall k \in \Agt,\ (i,j) \in E_k,\ i,j \neq 0 \label{mipc:mtz}\\
y_{ik} \leq t_{ik}, w_{ik} \leq \MT y_{ik} \quad \forall k \in \Agt,\ i \in \Aloc_k \label{mipc:eandl}\\
\EXPL_c \leq \displaystyle\sum_{k \in \Agt}\displaystyle\sum_{i \in \Aloc_k} \expl_k(i,c)w_{ik}\quad \forall c \in \Bloc \label{mipc:expl1}\\
0 \leq \EXPL_i \leq C^{0}_m(c)  \quad \forall c \in \Bloc \label{mipc:expl2} \\
t_{ik},\ w_{ik} \in \mathbb{N} \quad \forall k \in \Agt, i \in \Aloc_k  \label{mipc:ivar} \\
x_{ijk},\ y_{jk} \in \lbrace 0, 1 \rbrace \quad \forall k \in \Agt ,\ i,j \in \Aloc_k  \label{mipc:bvar}
\end{gather}  
}

\vspace*{-3.0ex}
\end{minipage}
\caption{MIP formulation of mission planning.\vspace*{-2ex}}
\label{milp:formulation}
\end{figure}

The MIP formulation for SAR planning is presented in
Fig.~\ref{milp:formulation}.  
A dummy vertex (denoted by $0$) has been introduced 
to represent the starting point and ending point of the agent paths.
Constraints
(\ref{mipc:ini}-\ref{mipc:end}) ensure that paths start and end 
at the dummy vertex $0$. 
Path continuity is guaranteed by (\ref{mipc:flow}).
Constraints (\ref{mipc:mtz}) eliminate subtours 
and, together with (\ref{mipc:eandl}),
they define the bounds of variables $t$ and $w$.
The coverage of each cell, as explained in (\ref{eq:expi}) is
bounded by (\ref{mipc:expl1}-\ref{mipc:expl2}). 
Finally,(\ref{mipc:ivar}-\ref{mipc:bvar}) set the
integer and binary requirements on the model variables.
This formulation represents the core of the model which can be used to maximize the 
coverage of joint
agent plans. Additional details can be found
in~\cite{AugFeo12:tr-idsia:05-12-n}.
\section{Proximity Directives}
In this section we extend the previous MIP model 
with a general class of proximity directives,
that is, direct dependencies in space and time among the plans of individual agents.
We consider the {\em physical distance} 
as the element to control through these directives.
The estimated distance, in meters, between any two agents $k,l \in \Agt$,
carrying out activities at sectors $i \in \Aloc_k,\ j \in \Aloc_l$
is denoted by $\ytwo_{ij}$.
Since sectors are environment's regions,
the value of $\ytwo_{ij}$ 
can be conveniently taken as the distance between sectors' centroids.


Given two disjoint subsets of agents $\Agt', \Agt'' \subset \Agt$,
we introduce two variables, the minimum distance between $\Agt'$ and $\Agt''$ at time $t$
denoted by $\MinDist_{\Agt', \Agt''}$, and conversely, the maximum distance, 
represented by $\MaxDist_{\Agt', \Agt''}$.
Let $p_t: \Agt \mapsto \Aloc$ be the {\em position} (i.e., assigned sector) 
of an agent at time step $t$ in 
the current solution. 
Using this notation, model variables $\MinDist$ and $\MaxDist$ are defined as follows:
\vspace*{-2ex}

{\small
\begin{gather}
\MinDist_{\Agt'\Agt''} = \min_{k \in \Agt',\ l \in \Agt'', i = p_t(k),\ j=p_t(l)} \ytwo_{ij} \label{mindist}\\
\MaxDist_{\Agt'\Agt''} = \max_{k \in \Agt',\ l \in \Agt'', i = p_t(k),\ j=p_t(l)} \ytwo_{ij} \label{maxdist} 
\end{gather}
}
\vspace*{-1ex}

\begin{figure}
\centering
\includegraphics[width=.50\columnwidth]{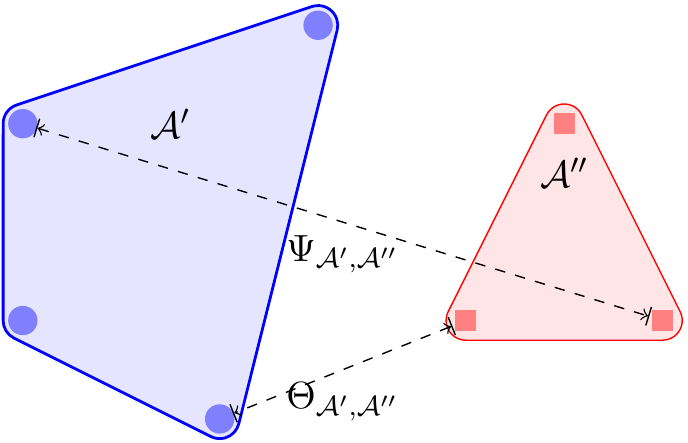}
\caption{Example of  the $\MinDist$ and $\MaxDist$  proximity variables. \vspace*{-4ex}}
\label{fig:proximity}
\end{figure}

Fig.~\ref{fig:proximity} graphically shows $\MinDist$ and $\MaxDist$ given the 
positions of two groups of agents (depicted in blue and red correspondingly) 
at an arbitrary point in time.
Thus, a proximity directive consists of 
defining desired lower or upper limits to
one of $\MinDist$ or $\MaxDist$ at a any single time step $t$.
We classify, and name, the proximity directives
according to the effect that can be achieved through their use.
The first type, called {\em coalition} directives, 
corresponds to those that set upper limits to the
maximum distance ($\MaxDist \leq $ ). 
Using these directives, two groups of agents can be kept
close to each other such as forming temporary coalitions.
Secondly, the {\em network} directives set upper limits to 
the minimum distance ($\MinDist \leq $). 
These constraints are typically employed to
enhance and promote communication (i.e., network connectivity) 
when data exchange in the
field is supported through the use of wireless ad hoc networks~\cite{FeoKudGamDic13:ssrr}.
%
Next, the {\em interference avoidance} directives, 
are those that keep two groups of agents
distant from each other such as to avoid task interference, 
dangerous situations (e.g., possibility of collisions),
and other undesirable events that are likely to occur when these agents get closer to each other.
These constraints are defined by setting lower limits to the minimum distance ($ \leq \MinDist$).
Finally, the {\em sparsity} directives, 
are those where the maximum distance should be greater than a certain value ($ \leq \MaxDist$).
Their effect is that of enlarging the area covered by members belonging to the corresponding groups.

The use of proximity directives represents a {\em strategic decision tool}.  They can be established
at specific points in time ({\em recurrent, periodic}, or {\em episodic}) or for the whole mission.
The directives can be enforced as {\em hard} constraints on the solutions, 
or their violation can be reflected as {\em penalties} in the objective function, which is the
approach we followed for the experiments showed in the following. 

Due to space limitations, the reader is referred to \cite{OctFeo13:tr-idsia:06-13-n} for
the derivation of the set of linear constraints that define (\ref{mindist}), (\ref{maxdist}), 
and each type of proximity directives.


\section{Evaluation}
\label{sec:evaluation}
In this section we show the use of the developed framework 
through a series of simulation experiments. 
We consider three different types of agents commonly used in
WiSAR missions, namely,
(i) aerial robotic platforms (UAVs), 
(ii) human rescuers,
and (iii) air-scent dogs.
We employ proximity directives 
to achieve  the realization of three different strategies aiming
to address three specific issues arising in WiSAR missions:
the reduction of {\em task interference}, the {\em promotion of synergies}, 
and the {\em enlargement of the operational area} of specific agents.

In the first application, 
we aim at reducing the task interference caused by the presence of
ground agents (i.e., human rescuers and other dogs) in the vicinity
of air-scent dogs. 
In the second one, we study the use of proximity directives to promote synergistic cooperation 
between human rescuers and UAVs.
For instance, in real deployments, 
if a human rescuer and a robot are concurrently exploring 
nearby or overlapping areas,
the robot could serve on-demand requests for the
real-time video/image streaming of areas selected by the human.
These on-site requests sent by the human can be employed to 
command the robot to inspect parts of the area 
that are difficult to reach to the human.
Finally, in the last set of experiments, we focus 
on defining mission plans that result in a widespread distribution of UAV units, i.e., a larger
operational area, which represents a way to guarantee their fast availability in different parts of
the field. This is based on the observation that UAVs can be an important asset in WiSAR missions,
and can be used to provide many different forms of support (e.g., for visual and communication
tasks).

For each application, we defined a set of proximity directives aiming at
promoting the corresponding strategy.
For solving the MIP 
we use the CPLEX\textregistered\  problem solver.  In order to speed up the computation of plans, 
the solver was enhanced with ad hoc heuristics  developed by the authors.

  \vspace*{-1ex}
\subsection{Scenarios}
 \vspace*{-1ex}

\begin{figure}
\centering
\subfloat[DEM]{\label{fig:dem}\includegraphics[width=.3\columnwidth]{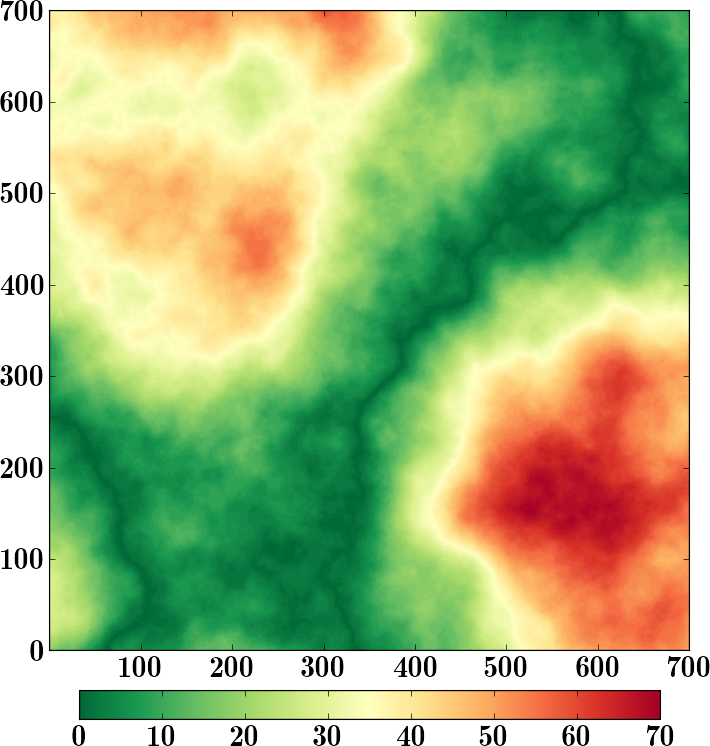}}
\subfloat[Vegetation]{\label{fig:veg}\includegraphics[width=.3\columnwidth]{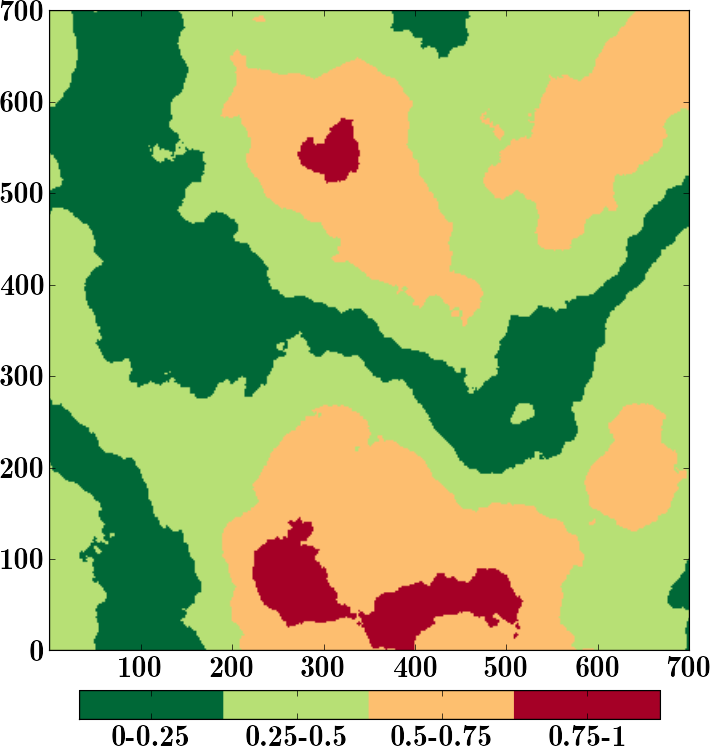}}
 \vspace*{-1ex}
\caption{Characteristics of the considered test area. \vspace*{-0ex}}
\label{fig:area}
\end{figure}

\begin{figure}
\centering
\includegraphics[width=0.74\columnwidth]{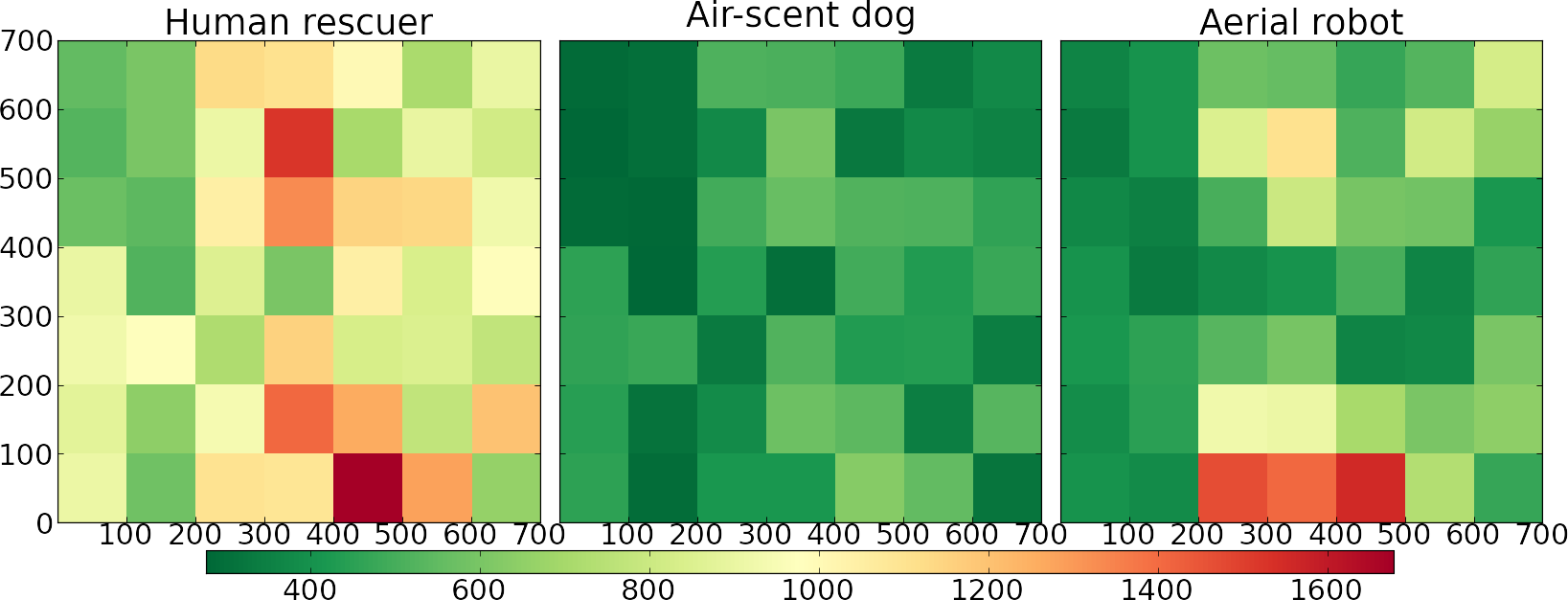}
\caption{Search efficacy as time to fully cover a cell. \vspace*{-0ex}}
\label{fig:search_efficacy}
\end{figure}

The test scenarios are based on a real-world area of size 700$\times$700 $m^2$.  
Figure~\ref{fig:area} shows the
digital elevation map (DEM) and the distribution of vegetation of the area.  
The area has been decomposed into cells of 100$\times$100 $m^2$.  
To account for the increased mobility of aerial robots compared to dogs and
humans, we define the size of their sectors as 200$\times$200 $m^2$ (i.e., all clusters are of 2$\times$2 cells). 
For human and air-scent dog agents, the sectors are of same size as the
cells.  
The traversability graphs allows the movement between contiguous sectors.
Figure~\ref{fig:search_efficacy} shows the way the
search efficacy was defined at each cell, for each type of agent, taking into
account the local geographical properties.
Plans are computed for a mission time-span of 35 minutes, 
with $\Delta_t = $ 300 seconds.  
A maximum time of one hour was given to the solver to
find the best possible solution.  
Within the allowed time, the 
{\em optimality gap} of the  obtained solutions was
$5\%$, on average.
For each computed mission plan, we ran 50 simulations using a
{\em discrete-time simulator
  for SAR scenarios}, derived from our previous work~\cite{FeoGamDic12:rosin, AugFeo12:tr-idsia:05-12-n}.
\vspace*{-2ex}
\subsection{Results}
\subsubsection{Interference avoidance}
At first, we demonstrate the use of {\em interference avoidance directives}.
We simulate the impact of interference inside the random mobility models in the following way.  At
all simulation steps, a dog becomes affected by interference according to a random rule: whenever
another potentially interferer agent (i.e., another dog or a human rescuer) falls within the
simulated interference range (100m in the experiments), a dog experience interference with a
probability $p=(100 - d)/ 100$.  If a random Bernoulli trial succeeds, then the dog will be
attracted towards interferer's position.  As soon as the dog arrives to that position, or if the
current task changes, the interference is canceled, and the dog continues with the execution of its
original plan.  

\begin{figure}
\centering
\includegraphics[width=1.0\columnwidth]{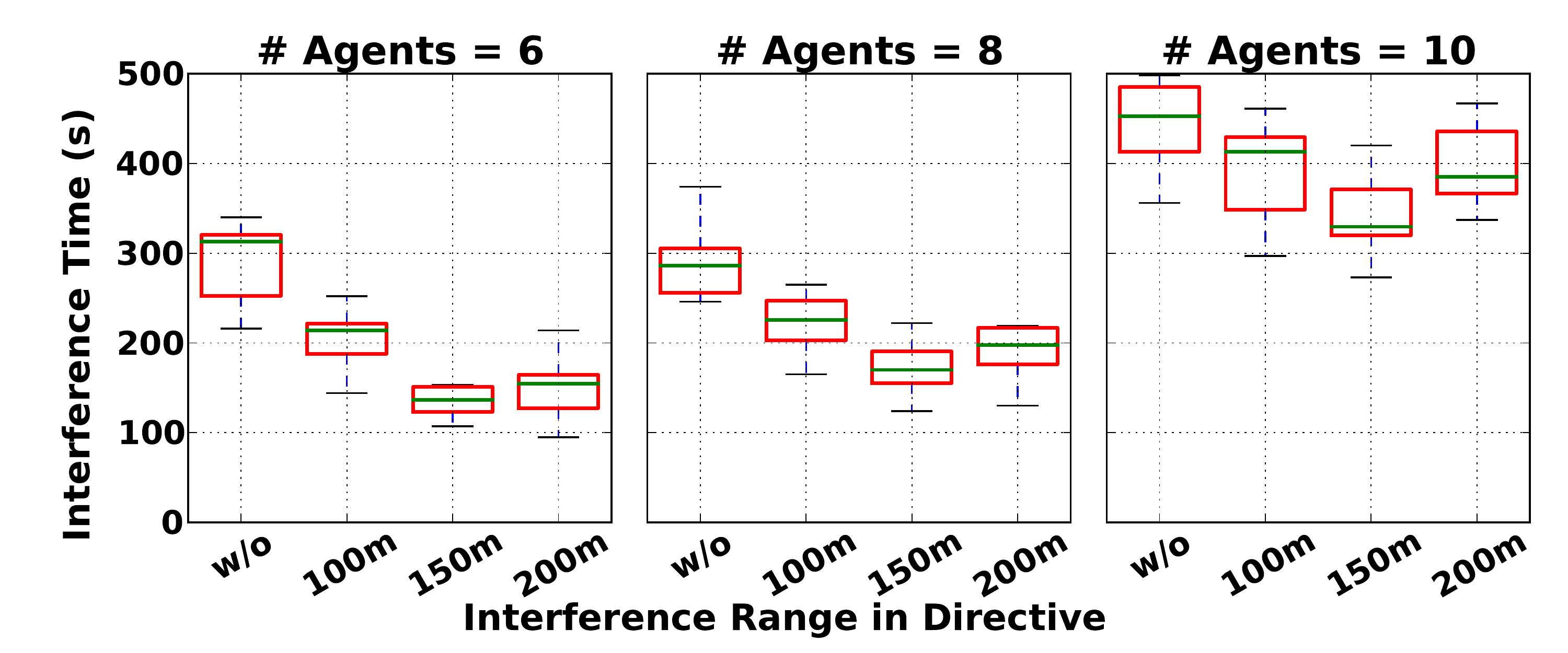}\\
\includegraphics[width=1.0\columnwidth]{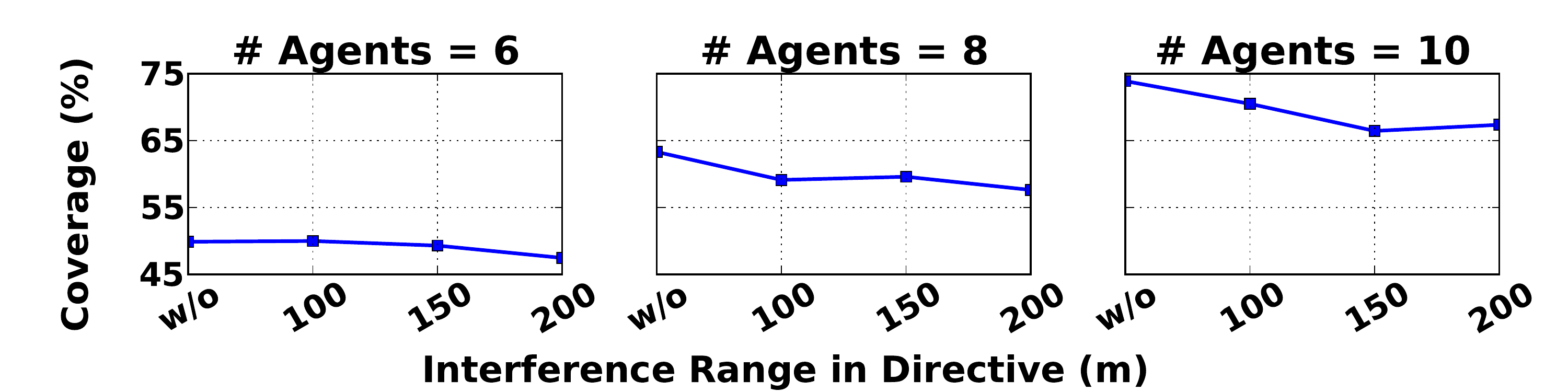} 
\caption{Effects of prevention of {\em task interference}.\vspace*{-4ex} }
\label{fig:res_interference}
\end{figure}

For testing the effect of using the directives to avoid interference, we considered three problem
instances, with teams composed by 6, 8, and 10 agents, with an equal number of human rescuers and
air-scent dogs. We compare the plans obtained with and without the use of the directives. In
particular, we considered three different sets of directives, with increasing interference ranges
(100m, 150m, and 200m). 
The results in Fig.~\ref{fig:res_interference} (upper figure) show the mean
interference time (in simulation) for all dogs in the team (y-axis), over the 
interference ranges considered (x-axis).
In the lower figure we can observe the resulting percentage of area coverage (y-axis), 
for each one of the computed mission plans.  
It can be observed that the use of the directives diminished the
effect of interference at a expense of a small reduction in area coverage.  
Moreover, as the interference range increases in the specification of the directive, 
the lesser the effects of interference are.

\subsubsection{Boosting cooperation}
\begin{figure}
\centering
\includegraphics[width=1.0\columnwidth]{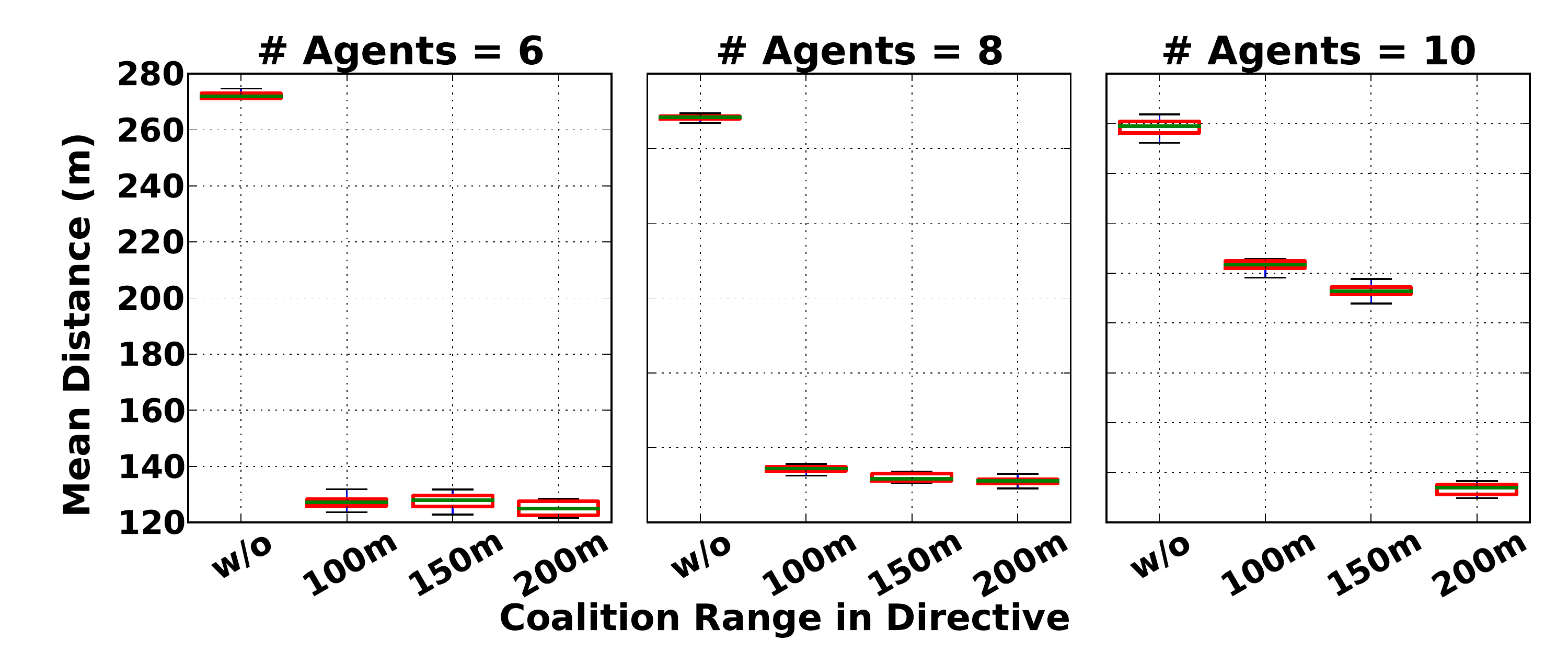}\\
\includegraphics[width=1.0\columnwidth]{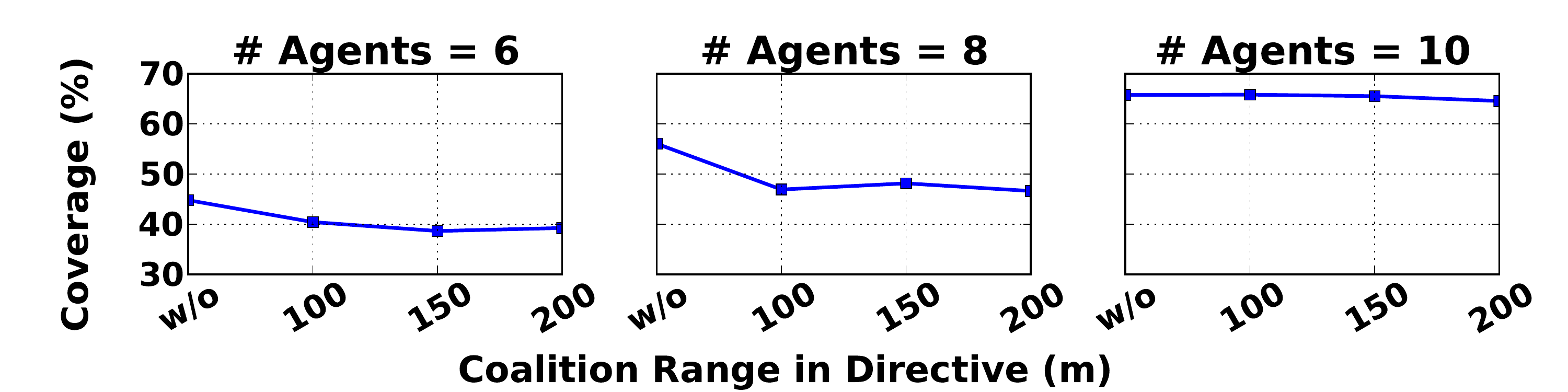}
\caption{Results for {\em promotion of cooperation}. \vspace*{-0ex}}
\label{fig:res_coalition}
\end{figure}

With the following set of experiments, we show the use of {\em coalition directives} to promote
cooperation between humans and robots.  The use of coalition directives is oriented at constraining
the maximum distance between a human and the group of robots, therefore increasing the likelihood of
having a robot in the vicinity of any human rescuer during the course of the mission.  
 We refer to the limit on the maximum distance as the {\em coalition range}.  
 We considered 3 problem instances, with teams composed by 6, 8, and 10 agents, 
 each with an equal number of human rescuers and robots.
Like in the previous test, we use three different sets of coalition directives, with increasing
coalition ranges (100m, 150m, and 200m).  
Results in Fig.~\ref{fig:res_coalition} (upper figure) show the mean
distance between humans and robots averaged over all the simulation time (y-axis) for
the different coalition ranges considered. 
In the lower figure, the resulting percentage of area coverage
for each of the computed mission plans.  
As we can appreciate, the use of directives decreases the
distances between robots and humans, 
at only a small expense of reduction in coverage.

\subsubsection{Enlargement of operational area}
\begin{figure}
\centering
\includegraphics[width=1.0\columnwidth]{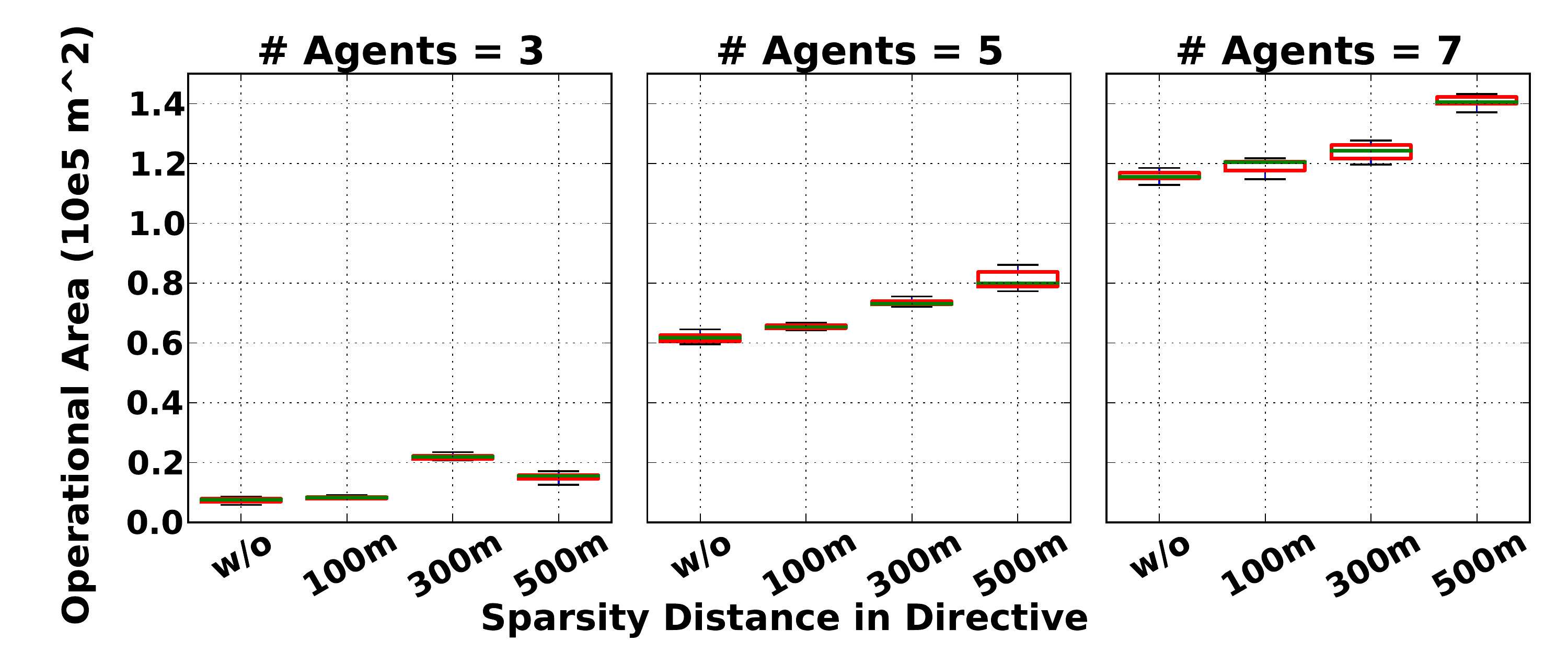}\\
\includegraphics[width=1.0\columnwidth]{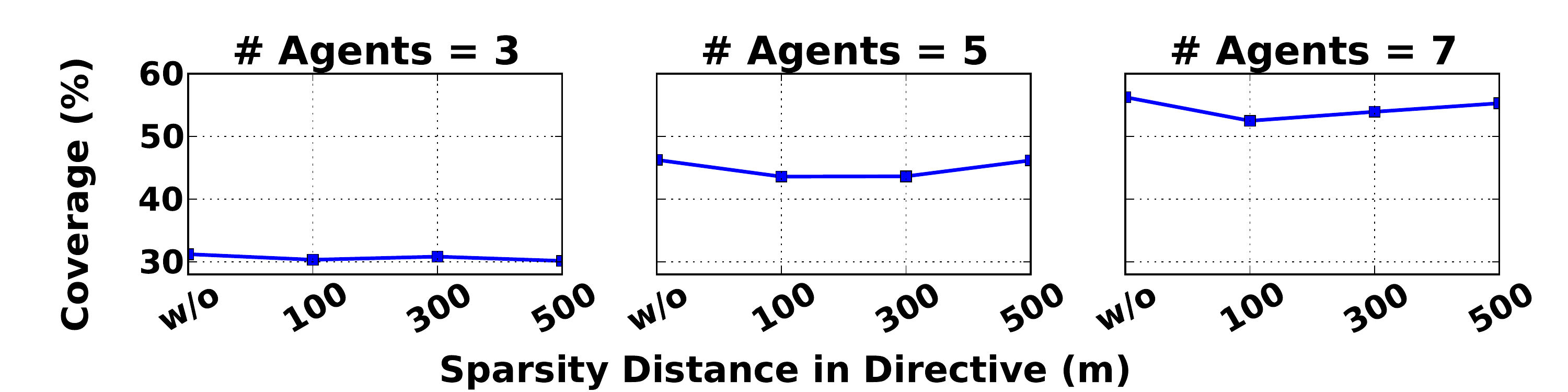}
\caption{Results of {\em enlargement of operational area}. \vspace*{-0ex}}
\label{fig:res_ocov}
\end{figure}

As final application, we make use of the 
{\em sparsity directives} to promote a widespread 
distribution of the UAVs over the mission area.
We considered 3 problem instances, with teams composed by 3, 5,
and 7 UAVs,
and use three different sets of directives, with
increasing sparsity distance (100m, 300m, and 500m).
As metric of sparsity we consider the area of the convex hull of
the positions of the robots projected on the surface (i.e., the operational area).
The results in Fig.~\ref{fig:res_ocov} (upper figure) show the average of the size
operational area in $m^2$ (y-axis) over the sparsity distances considered inside the directives. 
In the lower figure, the resulting percentage of area coverage obtained 
from each one of the computed mission plans.
It can be observed that the use of the directives provides a great enlargement of the operational
area, again, at the expense of a small reduction in area coverage.

\section{Conclusions}
We presented a mixed integer linear formulation for mission planning in heterogeneous teams of
physical agents. 
The aim of the mathematical formulation is to assign plans to the agents by
exploiting their specific characteristics in relation to the tasks, and promoting their synergies.
Together with the formulation we introduced a class of powerful 
mission directives that can be used to promote or enforce 
of spatio-temporal relations among the 
search trajectories of the agents.
We demonstrated the application of the proposed framework
to address common issues arising in real-world missions.
%
Future work will address the
definition of decentralized strategies, where each agent autonomously adapts its own plan while
coordinating with the others.

\section*{Acknowledgments}
This research has been partially funded by 
the Swiss National Science Foundation (SNSF) 
Sinergia project SWARMIX, project number {CRSI22\_133059}.
\bibliographystyle{IEEEtran}
\bibliography{swarmix}

\end{document}